\documentclass{article}

\usepackage[preprint,nonatbib]{neurips_2024}

\usepackage[utf8]{inputenc} 
\usepackage[T1]{fontenc}    
\usepackage{hyperref}       
\usepackage{url}            
\usepackage{booktabs}       
\usepackage{amsfonts}       
\usepackage{nicefrac}       
\usepackage{microtype}      
\usepackage{xcolor}         
\usepackage{multirow,multicol}

\usepackage{graphicx}
\usepackage{amsmath}
\usepackage{amssymb}
\usepackage{booktabs}

\usepackage{comment}
\usepackage{bm}
\usepackage{enumitem}
\usepackage{listings}
\usepackage{dblfloatfix}
\usepackage[labelfont=bf]{caption}

\usepackage{amsmath,amsfonts,bm, amssymb}
\usepackage{amsthm}
\usepackage{mathtools}
\usepackage{nicefrac}

\usepackage{setspace}
\usepackage{enumitem}
\usepackage[titles]{tocloft}

\newcommand{\appropto}{\mathrel{\vcenter{
  \offinterlineskip\halign{\hfil$##$\cr
    \propto\cr\noalign{\kern2pt}\sim\cr\noalign{\kern-2pt}}}}}
\DeclareMathOperator{\E}{\mathbb{E}}

\newcommand{\R}{\mathbb{R}}

\newcommand{\OurMethod}{Little-Big }

\usepackage{xcolor}

\definecolor{codegreen}{rgb}{0,0.6,0}
\definecolor{codegray}{rgb}{0.5,0.5,0.5}
\definecolor{codepurple}{rgb}{0.58,0,0.82}
\definecolor{backcolour}{rgb}{0.95,0.95,0.92}

\lstdefinestyle{mystyle}{
    backgroundcolor=\color{backcolour},   
    commentstyle=\color{codegreen},
    keywordstyle=\color{magenta},
    numberstyle=\tiny\color{codegray},
    stringstyle=\color{codepurple},
    basicstyle=\ttfamily\footnotesize,
    breakatwhitespace=false,         
    breaklines=true,                 
    captionpos=b,                    
    keepspaces=true,                 
    numbers=right,                    
    numbersep=5pt,                  
    showspaces=false,                
    showstringspaces=false,
    showtabs=false,                  
    tabsize=2
}

\title{Speeding Up Image Classifiers with Little Companions}

\author{
  Yang Liu \thanks{Corresponding author.}, \quad Kowshik Thopalli, \quad Jayaraman Thiagarajan  \\
  Lawrence Livermore National Laboratory\\
  \texttt{\{liu93, thopalli1, jayaramanthi1\}@llnl.gov} \\
}

\begin{document}

\maketitle

\begin{abstract}
  Scaling up neural networks has been a key recipe to the success of large language and vision models. However, in practice, up-scaled models can be disproportionately costly in terms of computations, providing only marginal improvements in performance; for example, EfficientViT-L3-384 achieves <2\% improvement on ImageNet-1K accuracy over the base L1-224 model, while requiring $14\times$ more multiply–accumulate operations (MACs). In this paper, we investigate scaling properties of popular families of neural networks for image classification, and find that scaled-up models mostly help with ``difficult'' samples. Decomposing the samples by difficulty, we develop a simple model-agnostic \textbf{two-pass} \OurMethod algorithm that first uses a light-weight ``little'' model to make predictions of all samples, and only passes the difficult ones for the ``big'' model to solve. Good little companions achieve drastic MACs reduction for a wide variety of model families and scales. Without loss of accuracy or modification of existing models, our \OurMethod models achieve MACs reductions of 76\% for EfficientViT-L3-384, 81\% for EfficientNet-B7-600, 71\% for DeiT3-L-384 on ImageNet-1K. \OurMethod also speeds up the InternImage-G-512 model by 62\% while achieving 90\% ImageNet-1K top-1 accuracy, serving both as a strong baseline and as a simple practical method for large model compression.
  
\end{abstract}

\begin{figure}[ht!]
    \centering

    \includegraphics[width=0.9\linewidth]{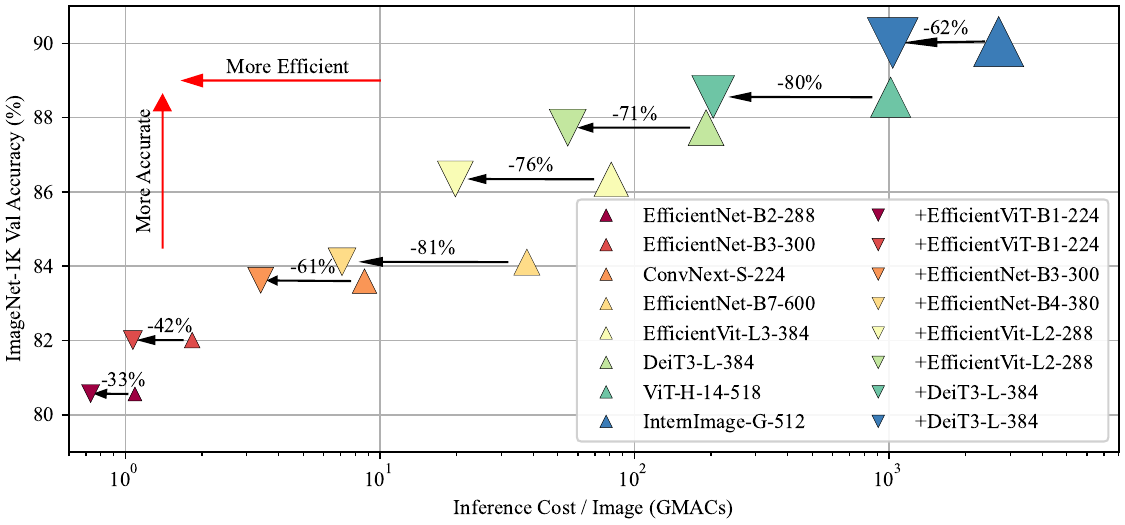}
    \vspace{-1mm}
    \caption{\OurMethod relaxes the assumption of obtaining predictions for samples in a single pass using a single model, achieving MACs reduction of $30\% - 80\%$ across models types (convolutional neural networks, transformers, and hybrid networks) and scales (from 1 to 3000 GMACs). Marker sizes correspond to $log(\# parameters)$. Model labels are formatted as ``Family-Size-InputResolution''.
    }
    \label{fig:examples}
    \vspace{-1mm}
\end{figure}
\clearpage
\section{Introduction}

Advances in parallel computing hardware, such as GPUs, have made end-to-end single-pass parallel processing standard in computer vision models. Large vision datasets like ImageNet-1K~\cite{imagenet} made it possible for such deep vision models (e.g.Alexnet~\cite{alexnet}, ResNet~\cite{resnet} and ViT~\cite{dosovitskiy2020vit})  to learn general visual features at scale. While vision models surpassed human performance on ImageNet-1K a decade ago~\cite{googlenet}, researchers are in the perpetual pursuit of achieving improved performance by using a combination of two strategies: 1) developing more performant models and training techniques for a given compute budget, and 2) scaling up the models.
While improved models from the former approach are often preferred in compute or memory-constrained applications, the latter has become increasingly popular, thanks to its success in large language models (LLMs)~\cite{touvron2023llama2}. However, despite architectural improvements, scaling up models remains expensive; we are often trading exponential compute cost for marginal gains in model accuracy (Table \ref{tab:scaling}). 

In this work, we show that much of the inefficiency comes from our implicit preference for \textbf{single-pass} models and propose an embarrassingly simple \textbf{two-pass} algorithm to drastically speed up models with little companions.

We summarize our work in response to two critical questions around model scaling and compression.

\begin{enumerate}[leftmargin=0.9cm,itemsep=0pt,topsep=0pt,label=\textbf{\textlangle Q\arabic*\textrangle}]
    \item \textit{Given a pair of \OurMethod models in the same model family, which incorrect predictions made by the Little model are fixed using the Big model?}\label{Q1}
\end{enumerate}

\begin{itemize}[leftmargin=*]
\item Using ImageNet-1K as the test bed and binning the Little model's predictions by confidence, we find that, very often, mistakes made by the Little model correspond to low confidence (measured via maximum softmax probability) 
\end{itemize}

\begin{enumerate}[leftmargin=0.9cm,itemsep=0pt,topsep=0pt,label=\textbf{\textlangle Q\arabic*\textrangle}]
    \setcounter{enumi}{1}
    \item \textit{Without compromising accuracy, can we speed up a Big model by using a Little model to preprocess a proportion of samples in a distribution? }\label{Q2} 
\end{enumerate}

\begin{itemize}[leftmargin=*]
\item We propose a two-pass Little-Big protocol where a light-weight \textbf{Little} model is used to make predictions (class and confidence) on all samples in the first pass, and a \textbf{Big} model performs a second pass on samples with low confidence from the first pass, achieving significant reduction in inference compute costs without compromising accuracy.

\end{itemize}

\begin{itemize}[leftmargin=*]
\item Without any modification to existing models, we prescribe \OurMethod pairs that significantly reduce compute costs of models across types and scales, while not compromising accuracy: \OurMethod models achieves MACs reductions of 76\% for EfficientViT-L3-384, 81\% for EfficientNet-B7-600, 71\% for DeiT3-L-384 on ImageNet-1K. \OurMethod also speeds up the very large InternImage-G-512 model by 62\% while achieving 90\% ImageNet-1K top-1 accuracy, serving both as a strong baseline and as a practical approach for efficient deployment of large models.
\end{itemize}

\begin{table*}[ht]
    \centering
        \caption{\textbf{Scaling up is expensive.} Scaling up model size is a popular way to improve performance without redesigning neural architectures or training recipes. Popular practices often involve sparing compound scaling in input resolution $H$, model width $w$ and depth $l$ over the base model (characterized by $H_0, w_0, l_0$). However Equation \ref{eq:model_scaling} shows that model size and inference cost quickly blow up by $10 - 100 \times$, with only marginal performance gains over small base models of the same family. }
    
    \resizebox{0.9\textwidth}{!}{
    
    \begin{tabular}{c c c c c c c c c}
        \toprule
          \multirow{2}{*}{Model Family} & \multirow{2}{*}{Size} & \multirow{2}{*}{$H/H_0$} & \multirow{2}{*}{$w/w_0$} & \multirow{2}{*}{$l/l_0$}& \multicolumn{2}{c}{ImageNet-1K Val} & \multicolumn{2}{c}{GMACs}\\
          \cmidrule(lr){6-7} \cmidrule(lr){8-9}
        
          & & & &  & Accuracy (\%) & $\Delta$ & Absolute & $\Delta$ \\
         
            \midrule

        \multirow{4}{*}{EfficientNet \cite{tan2019efficientnet}} & B0-224  & 1 & 1 & 1 &  $77.65$ & $--$ & $0.39$  & $--$ \\

        & B2-280 & 1.3 & 1.1 & 1.2 & $80.56$ & $+2.91$ & $1.09$  & $+1.8 \times$ \\

        & B4-380 & 1.7 & 1.4 & 1.8 & $83.45$ & $+5.80$ & $4.39$  & $+10.3\times$ \\

        & B7-600 & 2.7 & 2.0 & 3.1 & $84.11$ & $+6.45$ & $37.8$  & $+95.8\times$ \\
        
        \midrule

        \multirow{3}{*}{EfficientViT \cite{cai2023efficientvit}} & L1-224  & 1 & 1 & 1 & $84.39$ & $--$ & $5.3$  & $--$ \\
        
        & L2-288 & 1.3 & 1 & 1.4 & $85.60$ & $+1.21$ & $11.0$  & $+1.1\times$ \\
        
        & L3-384 & 1.7 & 2 & 1.4 & $86.34$ & $+1.95$ & $81.0$  & $+14.3\times$ \\
        
        \midrule
        
        \multirow{3}{*}{DeiT3 \cite{touvron2022deit3}} & S-224 & 1 & 1 & 1 & $83.05$ & $--$ & $4.6$  & $--$ \\
        
        & B-224 & 1 & 2 & 1 & $85.60$ & $+2.55$ & $17.6$  & $+1.1\times$ \\
        
        & L-384 & 1.7 & 2.7 & 2 & $87.73$ & $+4.68$ & $191.2$  & $+40.6\times$ \\
        \bottomrule
    \end{tabular}
    }
    \label{tab:scaling}
    \vspace{-3mm}
\end{table*}

\clearpage
\section{Related Work}

\subsection{Computer Vision Models}

Since Alexnet \cite{alexnet}, single-pass neural classifiers trained end-to-end have dominated leaderboards of various vision tasks from image classification to video segmentation \cite{googlenet,resnet,imagenet,beyer2020real,xu2018youtube}. Most neural models originates from two families of neural architectures: convolutional neural networks (CNNs) and transformers. Core to CNNs are ``convolutions'' which apply the same compute across locations on a feature map. On the other hand, transformers, which first found success in sequence-to-sequence language models \cite{vaswani2017attention} and subsequently in vision~\cite{dosovitskiy2020vit}, embed a sequence of tokens (e.g., image patches) and utilize attention mechanism to model intra- and inter-token interactions. Finally, hybrid models like Swin \cite{liu2021swin} combine CNN priors and attention mechanisms to achieve good performance.

\subsection{Scaling in Vision}

Many neural classifiers can be expressed as a composition of layers \cite{alexnet,googlenet,resnet,touvron2022deit3}:
\begin{equation}\label{eq:neural_classifier}
    y = F(x) = f_{w_l} \cdot ... \cdot f_{w_1} \cdot f_{w_0}(x), 
\end{equation}
where $x \in \R^{C \times H \times H}$ denotes an input image (using square images for simplicity) and $y \in (0,1)^{N}$ denotes a $N$-dimensional softmax confidence score. To get a class prediction, one finds the class $n$ with the highest confidence. $f_{w_j} $ denotes the function of layer $j$ with its characteristic width $w_j$. The inference cost of a single sample $x$ with such a model $F(x)$ can be expressed as:
\begin{equation}\label{eq:model_scaling}
    MACs[F(x)] \approx C_{F} * H^2 * w^2 * l 
\end{equation}
where $C_{F}$ is a scaling coefficient that is a function of the model family. In turn, the average inference cost per sample over a (finite) distribution $D$ is given by:
\begin{equation}\label{eq:macs}
    \E_D \bigg(MACs[F]\bigg) = \frac{1}{|D|} \sum_{x \in D} MACs[{F}(x)] \approx C_{F} * H^2 * w^2 * l
\end{equation}
Complementing innovations in model architecture $F$ that make models more compute efficient, a straightforward way of improving performance is to scale up the model. Thanks to architectural improvements like the skip connections \cite{resnet}, normalization layers \cite{googlenet}, as well as better initialization and parameterization\cite{yang2021tensor}, scaling up a model by orders of magnitude has been made feasible. Furthermore, efficient scaling strategies such as compound scaling in model width and depth, as well as input resolution \cite{tan2019efficientnet} have also emerged. However, Equation \ref{eq:model_scaling} imposes a fundamental limitation on the prohibitive cost of scaling: doubling in $H$, $w$, and $l$ leads to $2^5-1=31 \times$ increase in model compute cost. The marginal accuracy gains associated with model scaling shown in Table \ref{tab:scaling} further make it unappealing for many practical use cases with limited compute budget. 

\subsection{Model Compression} \label{sec:compression}

It is well known that modern neural networks have significant redundancy, thus impacting both the \textbf{model size} (often measured by the number of parameters) and inference \textbf{compute cost} (often measured by MACs). This has motivated the design of compression strategies to reduce redundancy in the model. Many lines of research focus on improving the inference time \cite{yin2022avit}, reducing both model size and compute simultaneously \cite{yu2023xpruner}, or trading model size for speed \cite{rao2021dynamicvit}.

Pruning is a popular way to reduce model compute cost by ablating weights that are deemed less essential for the prediction. For example, WDpruning \cite{yu2022wdpruning} reduces widths based on saliency scores, and X-pruner \cite{yu2023xpruner} measures a unit's importance by its contribution to predicting each target class. The implicit assumption is that the redundancy in size and compute are coupled, and one reduces compute by removing the units.

Another way to achieve model speedup is through adaptive computation mechanisms that ideally spend just enough compute on each sample. They often adopt some form of ``early exit'' mechanisms that reduce $l$ in Equation \ref{eq:model_scaling}. DynamicViT \cite{rao2021dynamicvit} downsamples the number of tokens adaptively to reduce compute cost, A-ViT \cite{yin2022avit} introduces learned token halting for ViT models so that not all tokens are processed by the full depth of the model, thus effectively reducing the average depth of the compute graph during inference.

\subsection{Human Vision}

While parallel processing plays an essential role in making it possible to ingest gigabits/s of raw visual information and compress it to tens of bits/s to guide our behavior, human vision is \textbf{not a single-pass process}. Human eyes have two distinct information processing pathways that originate from two types of photoreceptors called rods and cones. While the visual acuity in the cone-rich fovea is $\sim 1$ arcmin,  it only covers $\sim 2 $ degrees, or $\sim 0.01\%$ of the visual field \cite{rosenholtz2016capabilities}. The rest of the $\sim 99.9\%$ of the visual field is mostly dominated by rods which provides much lower visual acuity ($\sim 10$ arcmin). A given small patch in the visual field either only gets processed in a single-pass by the low-acuity rod pathway, or followed by additional passes with high-acuity foveal vision if needed as directed by the saccade. Studies \cite{tang2018recurrent,torralba2009pixels} have shown that human visual classification performance adapts to variable compute budget.
\section{Scaling Up Helps with ``Hard'' Samples}

\begin{figure}[h!]
    \centering
    \vspace{-3mm}
    \begin{tabular}{c}
          \includegraphics[width=0.95\linewidth]{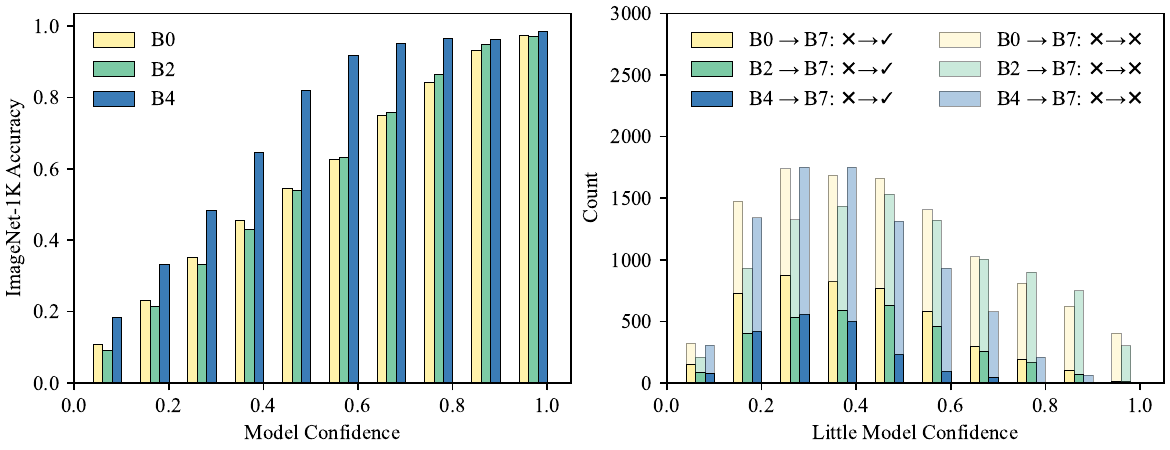}

    \end{tabular}
    \vspace{-1mm}
    \caption{Using the EfficientNet family as an example, we first show that confidence of individual models correlate well with prediction accuracy (\textbf{left}), which allows us to approximate a \textbf{``hardness''} axis with prediction confidence. Mistakes made by Little models can be categorized as \textit{correctable} (solid bars, $\times \rightarrow \checkmark$) and \textit{non-correctable} (shaded bars, $\times \rightarrow \times$) by the Big model. Breaking down the mistakes of Little models by hardness in the \textbf{right} panel, we find that very few ($<10\%$) correctable mistakes are easy (confidence > 0.5-0.7). This motivates the idea of a two-pass \OurMethod algorithm enabled by decomposing the samples by confidence thresholds. More examples can be found in Appendix Figure \ref{fig:confidence_appendix}.
    }
    \label{fig:confidence}
    \vspace{-1mm}
\end{figure}

In answer to \ref{Q1}, we first define an axis of ``hardness'' along which we can break down the predictions of the Little models. In lieu of an objective notion of hardness, we use the model confidence $max(F_{small}(x_{i}))$ as a surrogate since it reflects model calibration, i.e., higher prediction confidence correspond to higher accuracy (Figure \ref{fig:confidence} left).

Intuitively, there are 3 simple hypotheses on how scaled up ``Big'' models help correct the mistakes of base ``Little'' models: 

\begin{enumerate}[leftmargin=2.9cm,itemsep=0pt,topsep=0pt,label=\textbf{\textlangle H\arabic*\textrangle}]
    \item \textit{Big models uniformly help samples across difficulty,}\label{H1}
\end{enumerate}

\begin{enumerate}[leftmargin=2.9cm,itemsep=0pt,topsep=0pt,label=\textbf{\textlangle H\arabic*\textrangle}]
    \setcounter{enumi}{1}
    \item \textit{Big models preferentially help with ``hard'' samples,}\label{H2}
\end{enumerate}

\begin{enumerate}[leftmargin=2.9cm,itemsep=0pt,topsep=0pt,label=\textbf{\textlangle H\arabic*\textrangle}]
    \setcounter{enumi}{2}
    \item \textit{Big models preferentially help with ``easy'' samples.}\label{H3}
\end{enumerate}

Using the Pytorch \cite{pytorch} pretrained  EffcientNet family \cite{tan2019efficientnet} as an example, the right panel in Figure \ref{fig:confidence} visualizes prediction mistakes by Little model confidence (B0, B2, and B4) on the ImageNet-1K validation set. The mistakes made by Little models are divided into two categories, \textit{correctable} (solid bars) and \textit{non-correctable} (shaded bars) by the Big model. The full height of each bar (solid + shaded parts) sums up to the total number mistakes in the corresponding bin. 

Quantitatively, the average confidence of \textit{correctable} mistakes for EfficientNet-B0+B7, EfficientNet-B2+B7, EfficientNet-B4+B7 pairs are $0.38$, $0.41$, and $0.30$ respectively. In fact, $90\%$ of correctable mistakes fall under confidence thresholds of $0.65$, $0.67$, $0.47$, respectively. This suggests \ref{H2} is the likely to be true and motivates the two-pass algorithm in the next section.

\clearpage

\section{Two-Pass \OurMethod Algorithm}

\subsection{Speeding Up Big Models}

To answer \ref{Q2}, when we allow a sample to be solved with more than one pass like human vision, a simple way is to have a Little-Big pair $G_{F_{Little},F_{Big}}(x,T)$ or shortly $G(x,T)$ for simplicity:
\begin{equation}\label{eq:little_big}
    G_{F_{Little},F_{Big}}(x,T) = 
    \begin{cases}
        F_{Little}(x),  \text{if } max(F_{Little}(x)) \geq T \\
        F_{Big}(x). \\
    \end{cases}
    \qquad \text{\bf Little-Big}
\end{equation}

The essence of this Little-Big algorithm is to use a light-weight model to pre-screen samples and only pass hard samples to the Big model (see Algorithm). The average per-sample cost of inference over a dataset $D$ can then be expressed as:
\begin{align}\label{eq:MACs_little_big}
        \nonumber \E_D(MACs[G(x,T)]) & = \frac{1}{|D|} \bigg[\sum_{x \sim D} MACs[F_{Little}(x)] + \sum_{x \sim D^*} MACs[F_{Big}(x) ]\bigg] \\
        & = MACs[F_{Little}(x)] + \frac{|D^*|}{|D|} MACs[F_{Big}(x)],
\end{align}
where $D^* \subseteq D $ is defined as the set of $x$ where $max(F_{Little}(x)) < T, \forall x \in D$. 

\begin{figure}[h!]
    \centering

    \includegraphics[width=1.0\linewidth]{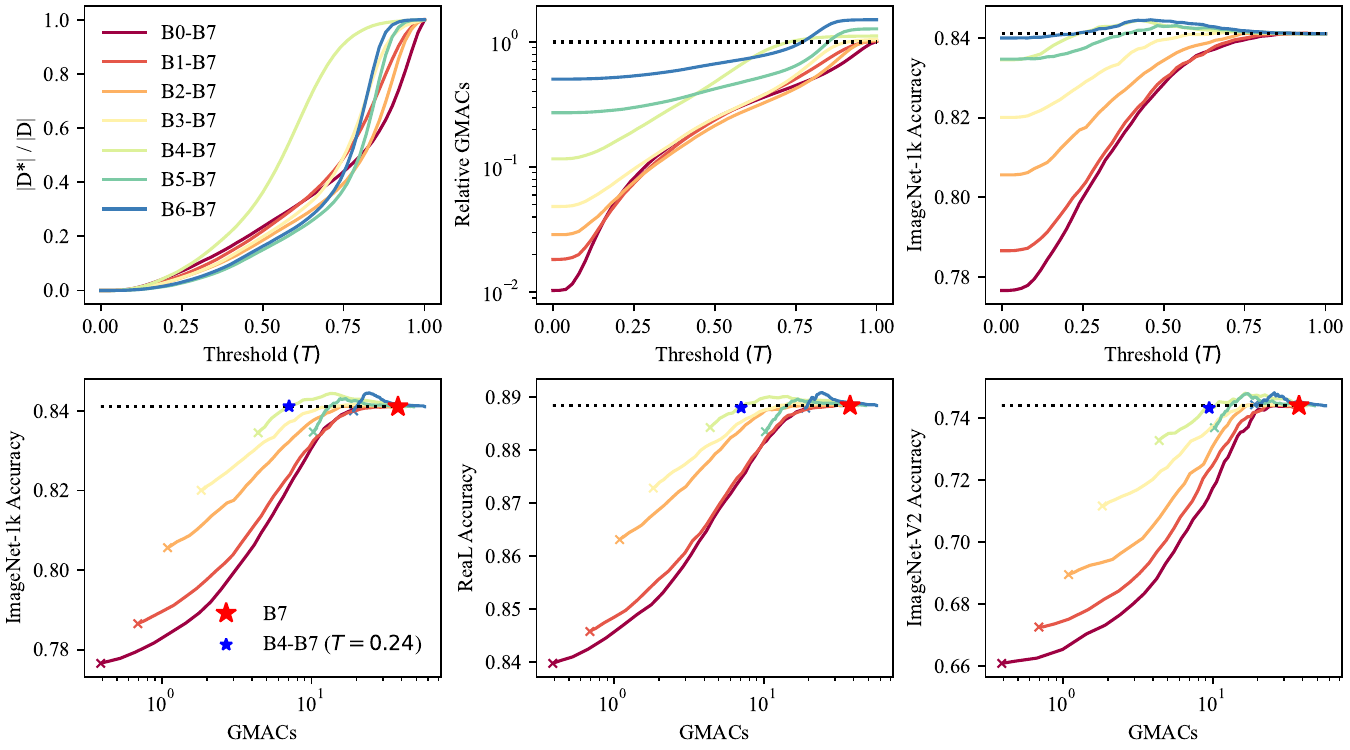}
    
    \caption{\textbf{Speeding up EfficientNet-B7 with smaller EfficientNets.} For using Little-Big, one needs to choose what small model to use and set a threshold $T$ based on an accuracy or MACs target. 50 evenly spaced $T$ in the range of 0 to 1 are sampled to generate each curve. ``\textcolor{red}{$\star$}'' marks the accuracy-MACs tradeoff of the Big EfficientNet-B7. ``\textcolor{blue}{$\star$}'' indicates the optimal Little-Big pair without any loss of accuracy on ImageNet-1K, achieving 81\% of MACs reduction. The same fixed $T$ performs well on both ReaL and ImageNet-V2 as well. Importantly, the optimal pair achieves both better accuracy and lower MACs than simply scaling down B7 to B6 or B5. 
    }
    \label{fig:efficientnet_b7}
    \vspace{-5mm}
\end{figure}

Using EfficientNet-B7 \cite{pytorch} as an example Big model to speed up, the top left panel in Figure \ref{fig:efficientnet_b7} shows how the relative size $|D^*|/|D|$ varies as a function of threshold $T$ with Little models ranging from EfficientNet-B0 to B6. The shape of the curves correspond to the cumulative distribution of prediction confidence for each Little model. Equation \ref{eq:MACs_little_big} further links $|D^*|/|D|$ to the relative compute cost $\E_D(MACs[G(x,T)])/\E_D(MACs[F_{Big}(x)])$: the higher the threshold the more samples that get passed to the big model, thus increasing MACs. Since $|D^*|/|D| \leq 1$, the upper bound of relative MACs overhead in the worst case scenario is $MACs[F_{Little}(x)]/ MACs[F_{Big}(x)]$, which is usually no greater than $1$ with proper choices of the Little model. As shown in the top middle panel of Figure \ref{fig:efficientnet_b7}, the net effect of adding a pre-screening Little model to the Big model leads to significant reduction in compute cost for a wide range of $T$ across difference choices of Little models.

\subsection{Speeding Up Without Losing Accuracy } \label{sec:choose_T}

\vspace{-2mm}
\begin{lstlisting}[language=Python, title={\textbf{\OurMethod Algorithm}: Pytorch pseudo code of \OurMethod for single image inference. Separate pre-processing image transforms are included as Big and Little models may require different input resolution and/or interpolation. The implementation keeps both models in the memory, Big and Little models can be loaded/unloaded to avoid overhead in max memory usage.}]
import torch.nn.functional as F

Class BigLittle:

    __init__(self, 
             little,   # small model
             big,      # big model
             t_little, # image transform for small model
             t_big,    # image transform for big model
             ):
             
        self.little, self.big   = little, big
        self.t_little, self.t_big = t_little, t_big
        
    predict(self,
            x,         # raw input image
            threshold  # prediction threshold T
            ):
            
        y = F.softmax(self.little(self.t_little(x)),dim=1)
        
        if torch.max(y) < threshold:
            y = F.softmax(self.big(self.t_big(x)),dim=1)
            
        return F.argmax(y,dim=1)

\end{lstlisting}
\vspace{-4mm}

\paragraph{Accuracy-MACs trade-off} In addition to MACs, one is interested in how the accuracy of $G(x,T)$ changes as a function of $T$. While it is not surprising that the accuracy generally decreases with smaller $T$, eventually degenerating to the accuracy of the Little model at $T=0$, certain choices of $G(x,T)$ achieve very little accuracy drop or even slight accuracy boost across a wide range of $T$. The bottom left panel of Figure \ref{fig:efficientnet_b7} visualizes the accuracy-MACs trade-off on ImageNet-1K. One might immediately notice that \textbf{any point} on the curves is a valid $G(x,T)$ model, and adjusting $T$ allows traversal along the full curve. Many points on certain curves, such as $G_{B4,B7}(x,T)$, achieves new Pareto optima. A convenient way to pick an optimal operating point of $G(x,T)$ is to set up a target accuracy either in the form of absolute accuracy or tolerable accuracy loss $\Delta Acc$. Setting $\Delta Acc >= 0$, one can easily draw a horizontal line on the accuracy-MACs plot (dashed line in bottom row of Figure \ref{fig:efficientnet_b7}) and find the leftmost intersection point (blue star, $G_{B4,B7}(x,T=0.24)$), yielding the Little-Big pair with the least compute cost while satisfying the accuracy target, speeding up B7 by $81\%$. 

\paragraph{Generalization beyond ImageNet-1K} To test whether the optimal $G$ determined on ImageNet-1K generalizes well, individual models as well as Little-Big pairs are evaluated two additional datasets: 1) ImageNet-ReaL \cite{beyer2020real} where the ground truth labels of the ImageNet-1K validation set are reassessed with an improved labeling protocol ((Figure \ref{fig:efficientnet_b7} bottom middle), and 2) ImageNet-V2 \cite{imagenetv2} where slightly ``harder'' validation images are collected (Figure \ref{fig:efficientnet_b7} bottom right). Comparing accuracy-MACs curves across the three datasets, one may notice that although the absolute accuracies and MACs varies, the shapes match qualitatively. Quantitatively, we find the optimal $G_{B4,B7}(x,T=0.24)$ found on ImageNet-1K performs well on ReaL and V2, with only marginal accuracy losses of $0.04\%$ and $0.07\%$, respectively, validating the generalizability of the optimal $G$.

Evidently, the key to determining the optimal $G$ is estimating the accuracy-MACs trade-off curves on the target distribution $D$. In practice, however, one might only have access to a small subset of $D$ or a set $D'$ close to to $D$. We simulate this case by determining an optimal $G(x,T)$ on the smaller V2 (10000 samples) and test generalization performance on ImageNet-1K and ReaL (50000 samples). Following the aforementioned process of determining optimal $G(x,T)$ on V2 (Appendix Figure \ref{fig:efficientnet_b7_appendix}) yields a similar optimal pair $G'_{B4,B7}(x,T=0.28)$), speeding up B7 by a similar $78\%$ on ImageNet-1K. This validates the robustness of such a process.

\clearpage

\subsection{Speeding Up Models Across Types And Scales}

\begin{table*}[h]
    \centering
        \caption{\textbf{The simple \OurMethod algorithm achieves strong MACs reduction across model types and scales.} Thresholds ($T$) are determined as the minimum value achieving a preset accuracy loss tolerance on ImageNet-1K (IN-1K).  All but InterImage-G-512 are compressed without loss in IN-1K accuracy. Models in \textcolor{blue}{blue} denote the Little model is from a different model family as the Big model. Configurations with lowest MACs are in \textbf{bold}. }
    
    \resizebox{0.99\textwidth}{!}{
    
    \begin{tabular}{c c c c c c c}
        \toprule
          \multirow{2}{*}{Model}& \multirow{2}{*}{Params} & \multicolumn{3}{c}{Top@1 Accuracy ($\%$) } & \multicolumn{2}{c}{GMACs}\\
          \cmidrule(lr){3-5} \cmidrule(lr){6-7}
        
           &  & IN-1k & ReaL & V2 & IN-1k\&ReaL & IN-V2 \\

        \midrule
        Efficientnet-B2-288 \cite{tan2019efficientnet} & $9.1M$ & $80.56$ & $86.31$ & $68.95$ & $1.09 $ & $1.09$\\

        +B0-224 $_{(T=0.66)}$ & $+3.5M_{(+58\%)}$ & $80.59_{(+0.03)}$ & $86.35_{(+0.04)}$ & $68.99_{(+0.04)}$ & $0.78_{(-28\%)}$ & $0.91_{(-16\%)}$\\

        \textcolor{blue}{+EfficientViT-B1-224} $_{(T=0.58)}$ & $+9.1M_{(+100\%)}$ & $80.57_{(+0.01)}$ & $86.35_{(+0.04)}$ & $68.92_{(-0.03)}$ & $\bm{0.73_{(-33\%)}}$ & $\bm{0.83_{(-24\%)}}$\\

        \midrule
        Efficientnet-B3-300 \cite{tan2019efficientnet} & $12.2M$ & $82.01$ & $87.28$ & $71.16$ & $1.83 $ & $1.83$\\

        +B1-240 $_{(T=0.66)}$ & $+7.8M_{(+64\%)}$ & $82.01_{(+0.00)}$ & $87.35_{(+0.07)}$ & $71.15_{(-0.01)}$ & $1.36_{(-26\%)}$ & $1.56_{(-15\%)}$\\

        \textcolor{blue}{+EfficientViT-B1-224} $_{(T=0.78)}$ & $+9.1M_{(+74\%)}$ & $82.01_{(+0.00)}$ & $87.31_{(+0.03)}$ & $70.94_{(-0.22)}$ & $\bm{1.07_{(-42\%)}}$ & $\bm{1.27_{(-31\%)}}$\\

        \midrule
        Efficientnet-B4-380 \cite{tan2019efficientnet} & $19.3M$ & $83.45$ & $88.43$  & $73.27$ & $4.6 $ & $4.6$ \\

        +B3-300 $_{(T=0.50)}$  & $+12.2M_{(+63\%)}$ & $83.46_{(+0.01)}$ & $88.42_{(-0.01)}$ & $73.39_{(+0.12)}$ & $2.7_{(-38\%)}$ & $\bm{3.1_{(-29\%)}}$ \\

        +B2-288 $_{(T=0.72)}$  & $+9.1M_{(+47\%)}$ & $83.47_{(+0.02)}$ & $88.42_{(-0.01)}$ & $73.28_{(+0.01)}$ & $\bm{2.7_{(-39\%)}}$ & $3.2_{(-27\%)}$ \\

        +B1-240 $_{(T=0.86)}$ & $+7.8M_{(+40\%)}$ & $83.46_{(+0.01)}$ & $88.44_{(+0.01)}$ & $73.27_{(+0.00)}$ & $4.0_{(-10\%)}$ & $4.3_{(-3\%)}$ \\

        +B0-224 $_{(T=0.94)}$  & $+5.3M_{(+27\%)}$ & $83.45_{(+0.00)}$ & $88.43_{(+0.00)}$ & $73.29_{(+0.02)}$ & $3.9_{(-10\%)}$ & $4.2_{(-4\%)}$ \\
        
        \midrule
        
        EfficientViT-B3-288 \cite{cai2023efficientvit} & $48.6M$ & $84.13$ & $88.49$  & $74.12$ & \multicolumn{2}{c}{$6.5$}\\

        +B3-224 $_{(T=0.60)}$ & $+48.6M_{(+100\%)}$ & $84.14_{(+0.01)}$ & $88.49_{(+0.00)}$  & $73.98_{(-0.14)}$ & $4.7_{(-28\%)}$ & $5.1_{(-22\%)}$\\
        
        +B2-288 $_{(T=0.76)}$ & $+24.3M_{(+50\%)}$ & $84.13_{(+0.00)}$ & $88.55_{(+0.06)}$ &  $73.82_{(-0.30)}$ & $\bm{3.8_{(-42\%)}}$ & $\bm{4.3_{(-33\%)}}$\\
        
        +B2-224 $_{(T=0.94)}$ & $+24.3M_{(+50\%)}$ & $84.14_{(+0.01)}$ & $88.52_{(+0.03)}$  & $74.11_{(-0.01)}$ & $\bm{3.8_{(-42\%)}}$ & $4.5_{(-30\%)}$\\
        \midrule
        ConvNext-L-224 \cite{liu2022convnext} & $197.8M$ & $84.39$ & $88.75$ & $74.34$ & \multicolumn{2}{c}{$34.3$}\\

        +S-224 $_{(T=0.52)}$ & $+50.2M_{(+25\%)}$ & $84.39_{(+0.00)}$ & $88.75_{(+0.00)}$ & $74.35_{(+0.01)}$ & $15.7_{(-54\%)}$ & $19.2_{(-44\%)}$\\
        
        \midrule
        Efficientnet-B7-600 \cite{tan2019efficientnet} & $66.3M$ & $84.11$ & $88.84$ & $74.39$ & \multicolumn{2}{c}{$37.8$}\\

        +B6-528 $_{(T=0.24)}$ & $+43.0M_{(+65\%)}$ & $84.13_{(+0.02)}$ & $88.90_{(+0.06)}$ & $74.50_{(+0.11)}$ & $20.1_{(-47\%)}$ & $21.4_{(-43\%)}$\\

        +B5-456 $_{(T=0.38)}$ & $+30.4M_{(+46\%)}$ & $84.12_{(+0.01)}$ & $88.78_{(-0.06)}$ & $74.65_{(+0.26)}$ & $13.2_{(-65\%)} $ & $15.4_{(-59\%)} $\\
        
        +B4-380 $_{(T=0.24)}$ & $+19.3M_{(+29\%)}$ & $84.12_{(+0.01)}$ & $88.80_{(-0.04)}$ & $74.32_{(-0.07)}$  & $\bm{7.1_{(-81\%)}} $ & $\bm{9.5_{(-75\%)}} $\\

        +B3-300 $_{(T=0.66)}$ & $+12.2M_{(+18\%)}$ & $84.13_{(+0.02)}$ & $88.88_{(+0.04)}$ & $74.41_{(+0.02)}$  & $14.6_{(-61\%)} $ & $18.9_{(-50\%)} $\\

        +B2-288 $_{(T=0.74)}$ & $+9.1M_{(+14\%)}$ & $84.11_{(+0.00)}$ & $88.83_{(-0.01)}$ & $74.39_{(+0.00)}$  & $15.6_{(-59\%)} $ & $20.1_{(-47\%)} $\\

        +B1-240 $_{(T=0.90)}$ & $+7.8M_{(+12\%)}$ & $84.12_{(+0.01)}$ & $88.85_{(+0.01)}$ & $74.34_{(-0.05)}$ & $32.9_{(-13\%)} $ & $34.6_{(-8\%)} $\\

        +B0-224 $_{(T=0.92)}$ & $+5.3M_{(+8\%)}$ & $84.12_{(+0.01)}$ & $88.84_{(+0.00)}$ & $74.40_{(+0.01)}$ & $28.4_{(-25\%)} $ & $31.1_{(-18\%)} $\\
        
        \midrule
        
        EfficientViT-L3-384 \cite{cai2023efficientvit} & $246.0M$ & $86.34$ & $89.66$  & $77.35$ & \multicolumn{2}{c}{$81.1 $} \\

        +L3-256 $_{(T=0.52)}$ & $+246.0M_{(+100\%)}$ & $86.35_{(+0.01)}$ & $89.71_{(+0.05)}$  & $77.36_{(+0.01)}$ & $40.4_{(-50\%)} $ & $44.3_{(-45\%)}$\\
        
        +L2-384 $_{(T=0.60)}$ & $+63.7M_{(+26\%)}$ & $86.34_{(+0.00)}$ & $89.83_{(+0.17)}$  & $77.55_{(+0.20)}$ & $27.0_{(-67\%)} $ & $31.7_{(-61\%)}$\\

        +L2-288 $_{(T=0.66)}$ & $+63.7M_{(+26\%)}$ & $86.35_{(+0.01)}$ & $89.86_{(+0.20)}$  & $77.37_{(+0.02)}$ & $\bm{19.8_{(-76\%)}} $ & $\bm{25.0_{(-69\%)}}$\\

        \midrule
        
        DeiT3-L-384 \cite{touvron2022deit3} & $304.8M$ & $87.73$ & $90.24$  & $79.34$ & \multicolumn{2}{c}{$191.2$} \\

        +B-224 $_{(T=0.82)}$ & $+86.6M_{(+28\%)}$ & $87.73_{(+0.00)}$ & $90.23_{(-0.01)}$  & $79.36_{(+0.02)}$ & $71.2_{(-63\%)} $ & $90.3_{(-53\%)}$\\

        \textcolor{blue}{+EfficientViT-L2-288} $_{(T=0.90)}$ & $+63.7M_{(+21\%)}$ & $87.73_{(+0.00)}$ & $90.37_{(+0.13)}$  & $79.43_{(+0.09)}$ & $\bm{54.7_{(-71\%)}} $ & $\bm{74.0_{(-61\%)}}$\\

        \midrule
        
        ViT-H-14-518 \cite{dosovitskiy2020vit} & $633.5$ & $88.55$ & $90.51$  & $81.12$ & \multicolumn{2}{c}{$1016$}\\

        +L-16-512 $_{(T=0.46)}$ & $+305M_{(+48\%)}$ & $88.59_{(+0.04)}$ & $90.89_{(+0.38)}$ & $81.04_{(-0.08)}$ & $430_{(-58\%)}$ & $488_{(-52\%)}$\\

        \textcolor{blue}{+DeiT3-L-384} $_{(T=0.70)}$ & $+305M_{(+10\%)}$ & $88.56_{(+0.01)}$ & $90.64_{(+0.13)}$ & $81.06_{(-0.06)}$ & $\bm{204_{(-80\%)}}$ & $\bm{276_{(-73\%)}}$\\
        
        \midrule
        
        InternImage-G-512 \cite{wang2022internimage} & $3.076B$ & $90.05$ & $90.97$  & $83.04$ & \multicolumn{2}{c}{$2700$} \\

        +XL-384 $_{(T=0.84)}$ & $+335M_{(+11\%)}$ & $90.01_{(-0.04)}$ & $90.98_{(+0.01)}$ & $82.99_{(-0.05)}$ & $1436_{(-47\%)}$ & $1781_{(-34\%)}$\\

        \textcolor{blue}{+DeiT3-L-384} $_{(T=0.90)}$ & $+305M_{(+10\%)}$ & $90.03_{(-0.02)}$ & $90.99_{(+0.02)}$ & $82.95_{(-0.09)}$ & $\bm{1035_{(-62\%)}}$ & $\bm{1335_{(-51\%)}}$\\
        
        \bottomrule
    \end{tabular}
    }
    \label{tab:main}
    
\end{table*}

Table \ref{tab:main} shows more examples the strong MACs compression with \OurMethod on a variety of model families including CNNs, transformers, and hybrid models, across scales from $1$ to $2700$ GMACs. Thanks to the model agnostic nature of \OurMethod, it allows pairing up models of different families (models in \textcolor{blue}{blue} in Table \ref{tab:main}). Large modes such as EfficientVit-L3-384 \cite{cai2023efficientvit}, DeiT3-L3-384 \cite{touvron2022deit3}, ViT-H-14-518 \cite{dosovitskiy2020vit} can be compressed without any loss of accuracy by $70\%$ to $80\%$.

It is worth noting that the roles of Big and Little in the \OurMethod pair are \textbf{relative}. For example, DeiT3-L-384 performs well as Little model, efficiently speeding up the 3B-parameter InternImage-G-512 \cite{wang2022internimage} and 600M-parameter ViT-H-14-518 by 62\% and 80\%, respectively. However, it does not imply that DeiT3-L-384 itself cannot act as the Big model and be sped up by an even smaller model. In fact, Table \ref{tab:main} shows that it can be sped up by the EfficientViT-L2-288 by 71\%. Similarly, EfficientNet-B4 can serve as a strong Little model for compressing EfficientNet-B7, as well as a Big model to be compressed by the smaller EfficientNet-B2. This enables the generalization of Little-Big to a $K$-pass framework where $K>1$ with additional performance gain (Appendix Section \ref{sec:exp_details}). Another important observation is that the speed up depends on the distribution $D$: Relative MACs reduction are consistently lower on the slightly harder ImageNet-V2 than on ImageNet-1K. This suggests that one may measure distribution shift by measuring average MACs with Little-Big.

\clearpage
\subsection{Comparison with Prior Art}

\begin{table*}[ht]
    \vspace{-5mm}
    \centering
        \caption{\textbf{\OurMethod outperforms a variety of pruning methods and adaptive compute models ({\textdagger})}. Models in \textcolor{blue}{blue} denote the Little model is from a different model family as the Big model. Thresholds in \textcolor{red}{red} indicate the Little model can simply replace the Big model, achieving model compression and better accuracy. ``+disl.'' denotes additional distillation training.}
    
    \resizebox{0.88\textwidth}{!}{
    
    \begin{tabular}{c c c c c c}
        \toprule
          \multirow{2}{*}{Model}& \multirow{2}{*}{Method} & \multicolumn{2}{c}{ImageNet-1K} & \multicolumn{2}{c}{GMACs}\\
          \cmidrule(lr){3-4} \cmidrule(lr){5-6}
        
           &  & Accuracy (\%) & $\Delta$ & Remaining & $\Delta$ \\
         
            \midrule

        \multirow{14}{*}{DeiT-S-224 \cite{touvron2021deit}} & Baseline \cite{yin2022avit} & $78.9$ & $--$ & $4.6$  & $--$ \\

        & DynamicViT-DeiT-S{\textdagger} \cite{rao2021dynamicvit} & $78.3$ & $-0.6$ & $3.4$  & $-26\%$ \\
        
        & A-ViT-S{\textdagger} \cite{yin2022avit} & $78.6$ & $-0.3$ & $3.6$  & $-22\%$ \\

        & A-ViT-S{\textdagger} + disl. \cite{yin2022avit} & $80.7$ & $+1.8$ & $3.6$  & $-22\%$ \\

        & eTPS-DeiT-S \cite{wei2023tps} & $79.7$ & $+0.8$ & $3.0$  & $-35\%$ \\

        & dTPS-DeiT-S \cite{wei2023tps} & $80.1$ & $+1.2$ & $3.0$  & $-35\%$ \\

        & SPViT-DeiT-S \cite{he2024attentiontoconv} & $78.3$ & $-0.6$ & $3.3$  & $-28\%$ \\

        & SPViT-DeiT-S + disl. \cite{he2024attentiontoconv} & $80.3$ & $+1.4$ & $3.3$  & $-28\%$ \\
        
        & \underline{Our Baseline} & $79.81$ & $+0.9$ & $4.6$  & $--$ \\
        
        & \textcolor{blue}{+EfficientViT-B1-224$_{(T=0.44)}$} & $79.81$ & $+0.9$ & $\bm{1.1}$  & $\bm{-77\%}$ \\
        
        & \textcolor{blue}{+EfficientNet-B2-288}\textcolor{red}{$_{(T=0.00)}$} & $80.56$ & $+1.7$ & $1.1$  & $-76\%$ \\

        & \underline{DeiT3-S Baseline} \cite{touvron2022deit3}  & $83.05$ & $+4.1$ & $4.6$  & $--$ \\

        & \textcolor{blue}{+EfficientViT-B1-224$_{(T=0.54)}$} & $\bm{83.11}$ & $\bm{+4.2}$ & $2.1$  & $-54\%$ \\
        
        & \textcolor{blue}{+EfficientNet-B2-288$_{(T=0.52)}$} & $83.06$ & $+4.1$ & $2.0$  & $-56\%$ \\
        
        \midrule
        
        \multirow{16}{*}{DeiT-B-224 \cite{touvron2021deit}} & Baseline & $81.81$ & $--$ & $17.6$  & $--$ \\

        & DynamicViT-DeiT-B{\textdagger} \cite{rao2021dynamicvit} & $81.4$ & $-0.4$ & $11.4$  & $-35\%$ \\
        
        & SCOP \cite{tang2020scop} & $79.7$ & $-2.1$ & $10.2$  & $-42\%$ \\

        & UVC \cite{yu2022uvc} & $80.57$ & $-1.24$ & $8.0$  & $-55\%$ \\

        & WDPruning \cite{yu2022wdpruning} & $80.76$ & $-1.05$ & $9.9$  & $-44\%$ \\

        & X-Pruner \cite{yu2023xpruner} & $81.02$ & $-0.99$ & $8.5$  & $-52\%$ \\

        & eTPS-DeiT-B \cite{wei2023tps} & $81.1$ & $-0.7$ & $11.4$  & $-35\%$ \\

        & dTPS-DeiT-B \cite{wei2023tps} & $81.2$ & $-0.6$ & $11.4$  & $-35\%$ \\

        & SPViT-DeiT-B \cite{he2024attentiontoconv} & $81.5$ & $-0.4$ & $8.4$  & $-52\%$ \\

        & SPViT-DeiT-B + disl. \cite{he2024attentiontoconv} & $82.4$ & $+0.6$ & $8.4$  & $-52\%$ \\

        & +DeiT-S $_{(T=0.40)}$ & $81.10$ & $-0.71$ & $6.5$  & $-63\%$ \\
        
        & +DeiT-S $_{(T=0.60)}$ & $81.83$ & $+0.02$ & $9.0$  & $-49\%$ \\

        & \underline{DeiT3-B Baseline} \cite{touvron2022deit3}  & $85.75$ & $+3.94$ & $17.6$  & $--$ \\

        &  +DeiT3-S \textcolor{red}{$_{(T=0.00)}$} & $83.05$ & $+1.24$ & $\bm{4.6}$  & $\bm{-74\%}$ \\
        
        & +DeiT3-S$_{(T=0.60)}$ & $85.75$ & $+3.94$ & $11.4$  & $-35\%$ \\

        & \textcolor{blue}{+EfficientViT-B2-288} $_{(T=0.60)}$ & $\bm{85.77}$ & $\bm{+3.96}$ & $7.8$  & $-56\%$ \\
        
        \midrule

        \multirow{9}{*}{Swin-S-224 \cite{liu2021swin}} & Baseline & $83.17$ & $--$ & $8.7$  & $--$ \\

        & STEP \cite{li2021step} & $79.6$ & $-3.6$ & $6.3$  & $-28\%$ \\
        
        & WDPruning \cite{yu2022wdpruning} & $81.8$ & $-1.4$ & $6.3$  & $-28\%$ \\
        
        & X-Pruner \cite{yu2023xpruner} & $82.0$ & $-1.2$ & $6.0$  & $-31\%$ \\

        & SPViT-Swin-S \cite{he2024attentiontoconv} & $82.4$ & $-0.6$ & $6.1$  & $-30\%$ \\

        & SPViT-Swin-S + disl. \cite{he2024attentiontoconv} & $83.0$ & $-0.2$ & $6.1$  & $-30\%$ \\
        
        & +Swin-T $_{(T=0.28)}$ & $82.03$ & $-1.14$ & $4.9$  & $-44\%$ \\
        
        & +Swin-T $_{(T=0.68)}$ & $\bm{83.18}$ & $\bm{+0.01}$ & $7.0$  & $-20\%$ \\

        & \textcolor{blue}{+EfficientViT-B2-224} $_{(T=0.58)}$ & $\bm{83.18}$ & $\bm{+0.01}$ & $\bm{2.5}$  & $\bm{-71\%}$ \\

        \bottomrule
    \end{tabular}
    }
    \label{tab:comparison}
\end{table*}

As discussed in Section \ref{sec:compression}, many methods have been developed for model compression in various axis. We compare accuracy-MACs tradeoff of \OurMethod with many such methods in Table \ref{tab:comparison}. Typical pruning methods such as WDPruning \cite{yu2022wdpruning} and X-Pruner \cite{yu2023xpruner} selectively removes units that are less important to model accuracy, methods like SPViT (2024) \cite{he2024attentiontoconv} ``prune'' some attention layers into convolutional layers, effecitvely changing the model type.  Addtionally, distillation is used to retrain the network to achieve better performance.  We show that even with tricks that effectively retrained models, many pruning methods are not competitive, especially compared with modern baselines such as DeiT3 \cite{touvron2022deit3}, which in essence are better trained ViTs. For example, the best performing SPViT-DeiT-B with distillation failed to outperform a better trained DeiT3-S baseline in both accuracy, model size, and MACs. It remains to be seen whether these methods work with the improved baseline models. In contrast, the process of choosing $T$ for lossless compression with Little-Big in Section \ref{sec:choose_T} will yield $T=0$ which automatically suggests replacement of the Big with the more performant Little model. Adaptive compute may still be an interesting direction, however popular models such as A-ViT \cite{yin2022avit} also fail to match the performance of better trained baseline models or show that their adaptive models are really better than simply scaling down the model moderately to match the MACs of the adaptive counterparts.

\section{Discussion}\label{sec:discussion}

A large corpus of literature in modern computer vision \cite{alexnet,googlenet,resnet,dosovitskiy2020vit,tan2019efficientnet,cai2023efficientvit,touvron2022deit3} has followed the norm of developing \textbf{single-pass} solutions trained \textbf{end-to-end} for a wide variety of vision tasks ranging from image/video classification to dense prediction. 

While many multi-pass test-time augmentations (TTAs) \cite{shanmugam2021bettertta} that aggregate predictions of several augmented views of the same sample have been developed as a post-hoc add-on to improve performance, these methods go in the opposite direction of \OurMethod. Fundamentally, $K$-pass TTAs are \textit{multiplicative} methods that trades $K \times $ inference cost for slight improvement of accuracy. \OurMethod is a \textit{subtractive} multi-pass algorithm that relies on a good decomposition of problems and solve each part with the least compute, not unlike Speculative Decoding in languange modelling \cite{leviathan2023speculative}. What TTA and \OurMethod share in common is that both are post-hoc methods that do not require any modification to the original models. It is actually possible to combine both in the same inference pipeline much like human vision to make predictions adaptively.

A potentially important contributing factor to the sub-optimal performance of popular adaptive compute models such as DynamicViT and A-ViT lies in their end-to-end training protocol. As the number of tokens decrease over depth, the deeper layers effectively ``see'' less pixels during training. This implicit coupling between the need to reduce compute cost at inference time and at training time due to the end-to-end training protocol may hinder the potential of adaptive methods. In contrast, \OurMethod decouples inference and training compute costs, and uses models individually trained on all pixels on full datasets.

\paragraph{Limitations} It is worth noting that, while \OurMethod achieves drastic MACs reduction across model families and scales, the speed up is \textbf{not free}: since an additional Little model is needed to make the \OurMethod pair, the storage overhead is usually a small fraction of the storage requirement of the Big model. However, the storage overhead does not necessarily translate to memory overhead. Although the example Pytorch pseudo code keeps both Big and Little models in memory at the same time to minimize latency while increasing maximum memory usage, one can alternatively only load one model into the memory at a time so that there is no overhead on maximum memory usage. Batching predictions can further reduce the memory I/O overhead per sample.

\paragraph{Extensions} It  possible to extend the same principles of \OurMethod to other tasks by re-examining and modifying Equation \ref{eq:neural_classifier}. For video classification,
where the input $x$ is updated as a sequence of images with an added time dimension $t$, $x \in \R^{C \times W \times H \times t}$. The inference cost of a neural video classifier is given by:
\vspace{-1mm}
\begin{equation}\label{eq:video_scaling}
    MACs[F(x)] \approx C_{F} * H^2 * t * w^2 * l, 
\end{equation}
which makes brute-force upscaling more costly because of the additional $t$ dimension. However, decomposition based on Equations \ref{eq:little_big} and \ref{eq:MACs_little_big} will most likely generalize to this task provided video classifiers are well calibrated and video samples are decomposable by some simple metrics like confidence. For dense image prediction tasks like semantic segmentation, the prediction $y$ becomes a map, where
$y \in [0,1]^{W \times H \times N}$.
In principle, one can continue perform decomposition on a per sample basis, but it may be more efficient to perform decomposition on a per pixel or per patch level. 
\vspace{-2mm}

\section{Conclusion}

In summary, we investigate how scaled-up models help base models correct their mistakes, show that the former preferentially help with samples with low confidence predictions. Inspired by that, we propose a simple two-pass \OurMethod algorithm that selectively pass ``difficult'' samples to large models, achieving drastic MACs reduction of up to $80\%$ without sacrificing accuracy for a wide range of model families and sizes.

The inefficiency in model scale-up and the effectiveness of \OurMethod in compressing the scaled-up models are two sides of the same coin, with effective decomposition of samples in a dataset being the key to connecting the two sides. Despite being embarrassingly simple and surprisingly effective, \OurMethod is considered as a lower bound on how much a model can be compressed without losing accuracy. More sophisticated ways to decompose the data distribution and better ways to use the Little models output are promising directions to further improve the performance of such subtractive multi-pass algorithms.

\section{Acknowledgement}

This work was performed under the auspices of the U.S. Department of Energy by the Lawrence Livermore National Laboratory under Contract No. DE-AC52-07NA27344. Supported by the LDRD Program under project 22-ERD-006. LLNL-JRNL-864664.

{\clearpage
\small
\bibliographystyle{ieee_fullname}
\bibliography{main}

\begin{thebibliography}{10}\itemsep=-1pt

\bibitem{beyer2020real}
Lucas Beyer, Olivier~J H{\'e}naff, Alexander Kolesnikov, Xiaohua Zhai, and A{\"a}ron van~den Oord.
\newblock Are we done with imagenet?
\newblock {\em arXiv preprint arXiv:2006.07159}, 2020.

\bibitem{cai2023efficientvit}
Han Cai, Junyan Li, Muyan Hu, Chuang Gan, and Song Han.
\newblock Efficientvit: Lightweight multi-scale attention for high-resolution dense prediction.
\newblock In {\em Proceedings of the IEEE/CVF International Conference on Computer Vision (ICCV)}, pages 17302--17313, October 2023.

\bibitem{imagenet}
Jia Deng, Wei Dong, Richard Socher, Li-Jia Li, Kai Li, and Li Fei-Fei.
\newblock Imagenet: A large-scale hierarchical image database.
\newblock In {\em 2009 IEEE conference on computer vision and pattern recognition}, pages 248--255. Ieee, 2009.

\bibitem{dosovitskiy2020vit}
Alexey Dosovitskiy, Lucas Beyer, Alexander Kolesnikov, Dirk Weissenborn, Xiaohua Zhai, Thomas Unterthiner, Mostafa Dehghani, Matthias Minderer, Georg Heigold, Sylvain Gelly, et~al.
\newblock An image is worth 16x16 words: Transformers for image recognition at scale.
\newblock {\em arXiv preprint arXiv:2010.11929}, 2020.

\bibitem{he2024attentiontoconv}
Haoyu He, Jianfei Cai, Jing Liu, Zizheng Pan, Jing Zhang, Dacheng Tao, and Bohan Zhuang.
\newblock Pruning self-attentions into convolutional layers in single path.
\newblock {\em IEEE Transactions on Pattern Analysis and Machine Intelligence}, 2024.

\bibitem{resnet}
Kaiming He, Xiangyu Zhang, Shaoqing Ren, and Jian Sun.
\newblock Deep residual learning for image recognition.
\newblock In {\em Proceedings of the IEEE conference on computer vision and pattern recognition}, pages 770--778, 2016.

\bibitem{alexnet}
Alex Krizhevsky, Ilya Sutskever, and Geoffrey~E Hinton.
\newblock Imagenet classification with deep convolutional neural networks.
\newblock {\em Advances in neural information processing systems}, 25, 2012.

\bibitem{leviathan2023speculative}
Yaniv Leviathan, Matan Kalman, and Yossi Matias.
\newblock Fast inference from transformers via speculative decoding.
\newblock In {\em International Conference on Machine Learning}, pages 19274--19286. PMLR, 2023.

\bibitem{li2021step}
Jiaoda Li, Ryan Cotterell, and Mrinmaya Sachan.
\newblock Differentiable subset pruning of transformer heads.
\newblock {\em Transactions of the Association for Computational Linguistics}, 9:1442--1459, 2021.

\bibitem{liu2021swin}
Ze Liu, Yutong Lin, Yue Cao, Han Hu, Yixuan Wei, Zheng Zhang, Stephen Lin, and Baining Guo.
\newblock Swin transformer: Hierarchical vision transformer using shifted windows.
\newblock In {\em Proceedings of the IEEE/CVF international conference on computer vision}, pages 10012--10022, 2021.

\bibitem{liu2022convnext}
Zhuang Liu, Hanzi Mao, Chao-Yuan Wu, Christoph Feichtenhofer, Trevor Darrell, and Saining Xie.
\newblock A convnet for the 2020s.
\newblock In {\em Proceedings of the IEEE/CVF conference on computer vision and pattern recognition}, pages 11976--11986, 2022.

\bibitem{cosine}
Ilya Loshchilov and Frank Hutter.
\newblock Sgdr: Stochastic gradient descent with warm restarts.
\newblock {\em arXiv preprint arXiv:1608.03983}, 2016.

\bibitem{loshchilov2017adamw}
Ilya Loshchilov and Frank Hutter.
\newblock Decoupled weight decay regularization.
\newblock {\em arXiv preprint arXiv:1711.05101}, 2017.

\bibitem{muller2021trivialaugment}
Samuel~G M{\"u}ller and Frank Hutter.
\newblock Trivialaugment: Tuning-free yet state-of-the-art data augmentation.
\newblock In {\em Proceedings of the IEEE/CVF International Conference on Computer Vision}, pages 774--782, 2021.

\bibitem{pytorch}
Adam Paszke, Sam Gross, Francisco Massa, Adam Lerer, James Bradbury, Gregory Chanan, Trevor Killeen, Zeming Lin, Natalia Gimelshein, Luca Antiga, Alban Desmaison, Andreas Kopf, Edward Yang, Zachary DeVito, Martin Raison, Alykhan Tejani, Sasank Chilamkurthy, Benoit Steiner, Lu Fang, Junjie Bai, and Soumith Chintala.
\newblock Pytorch: An imperative style, high-performance deep learning library.
\newblock In H. Wallach, H. Larochelle, A. Beygelzimer, F. d\textquotesingle Alch\'{e}-Buc, E. Fox, and R. Garnett, editors, {\em Advances in Neural Information Processing Systems 32}, pages 8024--8035. Curran Associates, Inc., 2019.

\bibitem{rao2021dynamicvit}
Yongming Rao, Wenliang Zhao, Benlin Liu, Jiwen Lu, Jie Zhou, and Cho-Jui Hsieh.
\newblock Dynamicvit: Efficient vision transformers with dynamic token sparsification.
\newblock {\em Advances in neural information processing systems}, 34:13937--13949, 2021.

\bibitem{imagenetv2}
Benjamin Recht, Rebecca Roelofs, Ludwig Schmidt, and Vaishaal Shankar.
\newblock Do imagenet classifiers generalize to imagenet?
\newblock In {\em International Conference on Machine Learning}, pages 5389--5400. PMLR, 2019.

\bibitem{rosenholtz2016capabilities}
Ruth Rosenholtz.
\newblock Capabilities and limitations of peripheral vision.
\newblock {\em Annual review of vision science}, 2:437--457, 2016.

\bibitem{shanmugam2021bettertta}
Divya Shanmugam, Davis Blalock, Guha Balakrishnan, and John Guttag.
\newblock Better aggregation in test-time augmentation.
\newblock In {\em Proceedings of the IEEE/CVF international conference on computer vision}, pages 1214--1223, 2021.

\bibitem{googlenet}
Christian Szegedy, Wei Liu, Yangqing Jia, Pierre Sermanet, Scott Reed, Dragomir Anguelov, Dumitru Erhan, Vincent Vanhoucke, and Andrew Rabinovich.
\newblock Going deeper with convolutions.
\newblock In {\em Proceedings of the IEEE conference on computer vision and pattern recognition}, pages 1--9, 2015.

\bibitem{tan2019efficientnet}
Mingxing Tan and Quoc Le.
\newblock Efficientnet: Rethinking model scaling for convolutional neural networks.
\newblock In {\em International conference on machine learning}, pages 6105--6114. PMLR, 2019.

\bibitem{tang2018recurrent}
Hanlin Tang, Martin Schrimpf, William Lotter, Charlotte Moerman, Ana Paredes, Josue~Ortega Caro, Walter Hardesty, David Cox, and Gabriel Kreiman.
\newblock Recurrent computations for visual pattern completion.
\newblock {\em Proceedings of the National Academy of Sciences}, 115(35):8835--8840, 2018.

\bibitem{tang2020scop}
Yehui Tang, Yunhe Wang, Yixing Xu, Dacheng Tao, Chunjing Xu, Chao Xu, and Chang Xu.
\newblock Scop: Scientific control for reliable neural network pruning.
\newblock {\em Advances in Neural Information Processing Systems}, 33:10936--10947, 2020.

\bibitem{torralba2009pixels}
Antonio Torralba.
\newblock How many pixels make an image?
\newblock {\em Visual neuroscience}, 26(1):123--131, 2009.

\bibitem{touvron2021deit}
Hugo Touvron, Matthieu Cord, Matthijs Douze, Francisco Massa, Alexandre Sablayrolles, and Herv{\'e} J{\'e}gou.
\newblock Training data-efficient image transformers \& distillation through attention.
\newblock In {\em International conference on machine learning}, pages 10347--10357. PMLR, 2021.

\bibitem{touvron2022deit3}
Hugo Touvron, Matthieu Cord, and Herv{\'e} J{\'e}gou.
\newblock Deit iii: Revenge of the vit.
\newblock In {\em European conference on computer vision}, pages 516--533. Springer, 2022.

\bibitem{touvron2023llama2}
Hugo Touvron, Louis Martin, Kevin Stone, Peter Albert, Amjad Almahairi, Yasmine Babaei, Nikolay Bashlykov, Soumya Batra, Prajjwal Bhargava, Shruti Bhosale, et~al.
\newblock Llama 2: Open foundation and fine-tuned chat models.
\newblock {\em arXiv preprint arXiv:2307.09288}, 2023.

\bibitem{vaswani2017attention}
Ashish Vaswani, Noam Shazeer, Niki Parmar, Jakob Uszkoreit, Llion Jones, Aidan~N Gomez, {\L}ukasz Kaiser, and Illia Polosukhin.
\newblock Attention is all you need.
\newblock {\em Advances in neural information processing systems}, 30, 2017.

\bibitem{internimage_repo}
Wenhai Wang, Jifeng Dai, Zhe Chen, Zhenhang Huang, Zhiqi Li, Xizhou Zhu, Xiaowei Hu, Tong Lu, Lewei Lu, Hongsheng Li, et~al.
\newblock Internimage.
\newblock \url{https://github.com/OpenGVLab/InternImage}, 2022.

\bibitem{wang2022internimage}
Wenhai Wang, Jifeng Dai, Zhe Chen, Zhenhang Huang, Zhiqi Li, Xizhou Zhu, Xiaowei Hu, Tong Lu, Lewei Lu, Hongsheng Li, et~al.
\newblock Internimage: Exploring large-scale vision foundation models with deformable convolutions.
\newblock {\em arXiv preprint arXiv:2211.05778}, 2022.

\bibitem{wei2023tps}
Siyuan Wei, Tianzhu Ye, Shen Zhang, Yao Tang, and Jiajun Liang.
\newblock Joint token pruning and squeezing towards more aggressive compression of vision transformers.
\newblock In {\em Proceedings of the IEEE/CVF Conference on Computer Vision and Pattern Recognition}, pages 2092--2101, 2023.

\bibitem{rw2019timm}
Ross Wightman.
\newblock Pytorch image models.
\newblock \url{https://github.com/rwightman/pytorch-image-models}, 2019.

\bibitem{xu2018youtube}
Ning Xu, Linjie Yang, Yuchen Fan, Dingcheng Yue, Yuchen Liang, Jianchao Yang, and Thomas Huang.
\newblock Youtube-vos: A large-scale video object segmentation benchmark.
\newblock {\em arXiv preprint arXiv:1809.03327}, 2018.

\bibitem{yang2021tensor}
Greg Yang and Edward~J Hu.
\newblock Tensor programs iv: Feature learning in infinite-width neural networks.
\newblock In {\em International Conference on Machine Learning}, pages 11727--11737. PMLR, 2021.

\bibitem{yin2022avit}
Hongxu Yin, Arash Vahdat, Jose~M Alvarez, Arun Mallya, Jan Kautz, and Pavlo Molchanov.
\newblock A-vit: Adaptive tokens for efficient vision transformer.
\newblock In {\em Proceedings of the IEEE/CVF Conference on Computer Vision and Pattern Recognition}, pages 10809--10818, 2022.

\bibitem{yu2022wdpruning}
Fang Yu, Kun Huang, Meng Wang, Yuan Cheng, Wei Chu, and Li Cui.
\newblock Width \& depth pruning for vision transformers.
\newblock In {\em Proceedings of the AAAI Conference on Artificial Intelligence}, volume~36, pages 3143--3151, 2022.

\bibitem{yu2023xpruner}
Lu Yu and Wei Xiang.
\newblock X-pruner: explainable pruning for vision transformers.
\newblock In {\em Proceedings of the IEEE/CVF Conference on Computer Vision and Pattern Recognition}, pages 24355--24363, 2023.

\bibitem{yu2022uvc}
Shixing Yu, Tianlong Chen, Jiayi Shen, Huan Yuan, Jianchao Tan, Sen Yang, Ji Liu, and Zhangyang Wang.
\newblock Unified visual transformer compression.
\newblock {\em arXiv preprint arXiv:2203.08243}, 2022.

\end{thebibliography}
}

\clearpage
\appendix

\section{Appendix}

\subsection{Additional Experimental Details} \label{sec:exp_details}

\textbf{Implementation} The three datasets used in this work, ImageNet-1K, 
ImageNet-ReaL, and ImageNet-V2 are released with ImageNet license \url{https://www.image-net.org/download.php},  Apache-2.0 license, and MIT license, respectively. All models are implemented in Pytorch \cite{pytorch}. EfficientNet \cite{tan2019efficientnet}, Swin \cite{liu2021swin}, ConvNext \cite{liu2022convnext}, and ViT \cite{dosovitskiy2020vit} checkpoints are loaded from public Torchvision pretrained models with BSD-3 license. Pretrained models including EfficientViT \cite{cai2023efficientvit}, DeiT \cite{touvron2021deit}, DeiT3 \cite{touvron2022deit3} are accessed on Timm \cite{rw2019timm} under Apache-2.0 license. The official code and weight of InternImage models \cite{internimage_repo} are accessed under MIT licence.

Model inference are conducted on an NVIDIA RTX 3090 with 24 GB vRAM, taking ~1 minute to ~4 hours to finsh evaluation on the ImageNet-1K validation set. The main results are based on measurements of single pretrained models. To estimate the error bar on single model accuracy, we train a small EfficientNet-B0 on ImageNet-1K from scratch for 300 epochs on 32 NVIDIA V100 GPUs with cosine learning rate decay \cite{cosine}, TrivialAugment \cite{muller2021trivialaugment}, input resolution of 224, batch size of 2048, AdamW \cite{loshchilov2017adamw} with initial learning rate of 0.003, and weight decay of 0.05. Three models are independently trained from scratch and achieve top-1 accuracies of $76.76\%$, $76.43\%$, $76.56\%$, with a standard deviation of $0.14 \%$. 

As shown in Section \ref{sec:choose_T}, the dataset used in choosing $T$ can affect the optimal $T$ as well as relative MACs reduction. By choosing $T$ over ImageNet-1K, Real, V2 for the same model pair, we estimate MACs reduction reported in Tables \ref{tab:main} and \ref{tab:comparison} to be $\sim 2\%$, which is an order of magnitude smaller than the effect size of $50\%$ throughout the paper, validating the statistical significance of our speedup.

\begin{table*}[ht]
    \vspace{-2mm}
    \centering
        \caption{To simulate errors stemming from noise in determining the optimal $T$, we choose $T$ for each Little-Big pair on ImageNet-1K, Real, and V2 and compute mean and standard deviation of accuracy change and MACs reduction on ImageNet-1K. }
    
    \resizebox{0.98\textwidth}{!}{
    
    \begin{tabular}{c c | c c | c c | c c | c c | c c }
        \toprule
          \multicolumn{2}{c}{Choose $T$ on: } & 
          \multicolumn{2}{c}{IN-1K} & 
          \multicolumn{2}{c}{ReaL} & 
          \multicolumn{2}{c}{$V2$} & 
          \multicolumn{2}{c}{ } & 
          \multicolumn{2}{c}{ }\\
          
        \midrule
           &  & 
           $\Delta Acc_{1K}$ & $\Delta GMACs_{1K}$ &
           $\Delta Acc_{1K}$ & $\Delta GMACs_{1K}$ & 
           $\Delta Acc_{1K}$ & $\Delta GMACs_{IN}$ & 
           $\Delta Acc_{1K}$ & $\Delta GMACs_{IN}$ &
           $\Delta Acc_{1K}$ & $\Delta GMACs_{IN}$\\
         
        \midrule

        Big Model & Little Model & &  & & & & & 
          \multicolumn{2}{c}{Mean} & 
          \multicolumn{2}{c}{SD}\\
        \midrule
        \multirow{7}{*}{EfficientNet-B7-600} & B6-528  & 
        $+0.02$ & $-47\%$ & 
        $-0.01$ & $-48\%$ & 
        $-0.11$ & $-49\%$ & 
        $-0.03$ & $-48\%$ & 
        $0.06$ & $1\%$\\

        & B5-456 & 
        $+0.01$ & $-65\%$ & 
        $+0.09$ & $-62\%$ & 
        $-0.20$ & $-68\%$ & 
        $-0.03$ & $-65\%$ & 
        $0.12$ & $2\%$\\
        
        & B4-380 & 
        $+0.01$ & $-81\%$ & 
        $+0.05$ & $-80\%$ & 
        $+0.09$ & $-78\%$ & 
        $+0.05$ & $-80\%$ & 
        $0.03$ & $1\%$\\

        & B3-300 & 
        $+0.02$ & $-61\%$ & 
        $-0.02$ & $-65\%$ & 
        $-0.01$ & $-69\%$ & 
        $-0.00$ & $-65\%$ & 
        $0.02$ & $3\%$\\

        & B2-288 & 
        $+0.00$ & $-59\%$ & 
        $+0.01$ & $-53\%$ & 
        $-0.16$ & $-67\%$ & 
        $-0.05$ & $-60\%$ & 
        $0.08$ & $6\%$\\

        & B1-240 & 
        $+0.01$ & $-13\%$ & 
        $-0.01$ & $-18\%$ & 
        $-0.12$ & $-51\%$ & 
        $-0.04$ & $-27\%$ & 
        $0.06$ & $17\%$\\

        & B0-224 & 
        $+0.01$ & $-25\%$ & 
        $+0.01$ & $-25\%$ & 
        $-0.03$ & $-44\%$ & 
        $-0.01$ & $-31\%$ & 
        $0.02$ & $9\%$\\

        \bottomrule
    \end{tabular}
    }
    \label{tab:error}
    
\end{table*}

As discussed in Section \ref{sec:compression}, the roles of Big and Little in the \OurMethod pair are \textbf{relative}. For example, DeiT3-L-384 performs well as Little model, efficiently speeding up the 3B-parameter InternImage-G-512 \cite{wang2022internimage} and 600M-parameter ViT-H-14-518 by 62\% and 80\%, respectively. However, it does not imply that DeiT3-L-384 itself cannot act as the Big model and be sped up by an even smaller model. In fact, Table \ref{tab:main} shows that it can be sped up by the EfficientViT-L2-288 by 71\%. Similarly, EfficientNet-B4 can serve as a strong Little model for compressing EfficientNet-B7, as well as a Big model to be compressed by the smaller EfficientNet-B2. This enables the generalization of Little-Big to a $K$-pass framework where $K>1$ with additional performance gain.

One can extend Little-Big to a 3-pass Tiny-Little-Big model $G_{F_{Tiny},F_{Little},F_{Big}}(x,T_1,T_2)$ or shortly $G(x,T_1, T_2)$ for simplicity:

\begin{equation}\label{eq:tiny_little_big}
    G_{F_{Tiny}, F_{Little},F_{Big}}(x,T_1, T_2) = 
    \begin{cases}
        F_{Tiny}(x),  \text{if } max(F_{Tiny}(x)) \geq T_1 \\
        F_{Little}(x),  \text{if } max(F_{Tiny}(x)) < T_1, max(F_{Little}(x)) \geq T_2\\
        F_{Big}(x). \\
    \end{cases}
\end{equation}

A 3-pass model $G_{B2,B4,B7}(x,T_1=0.74,T_2=0.26)$) further compresses EfficientNet-B7, achieving 85\% MACs reduction while preserving the Big model performance.

\clearpage
\subsection{Additional Data}

\begin{figure}[h!]
    \centering

    \begin{tabular}{c}
          \includegraphics[width=0.9\linewidth]{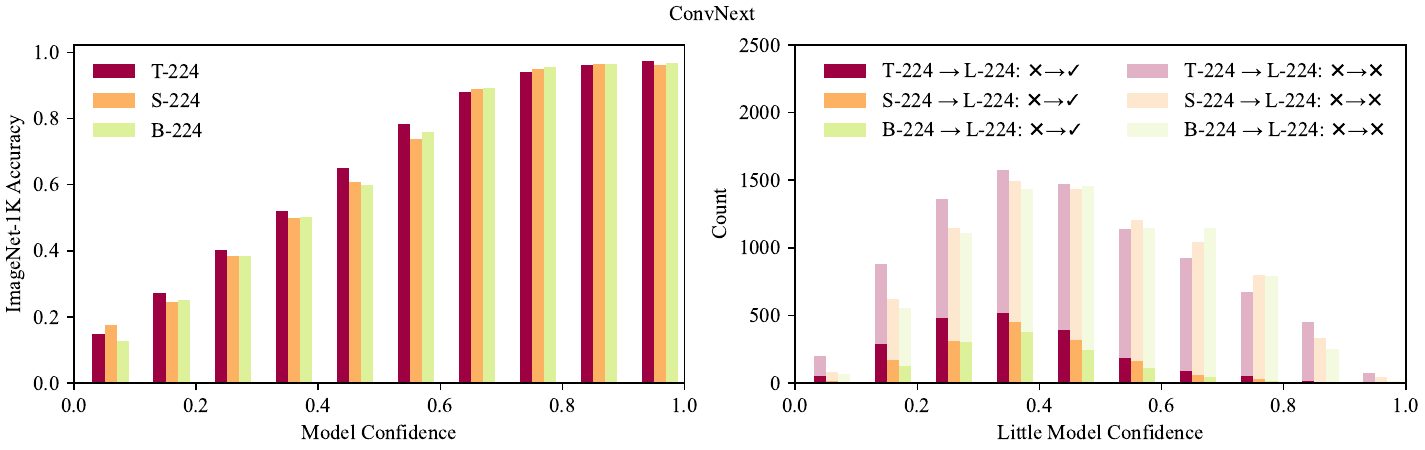}
          \\
          \includegraphics[width=0.9\linewidth]{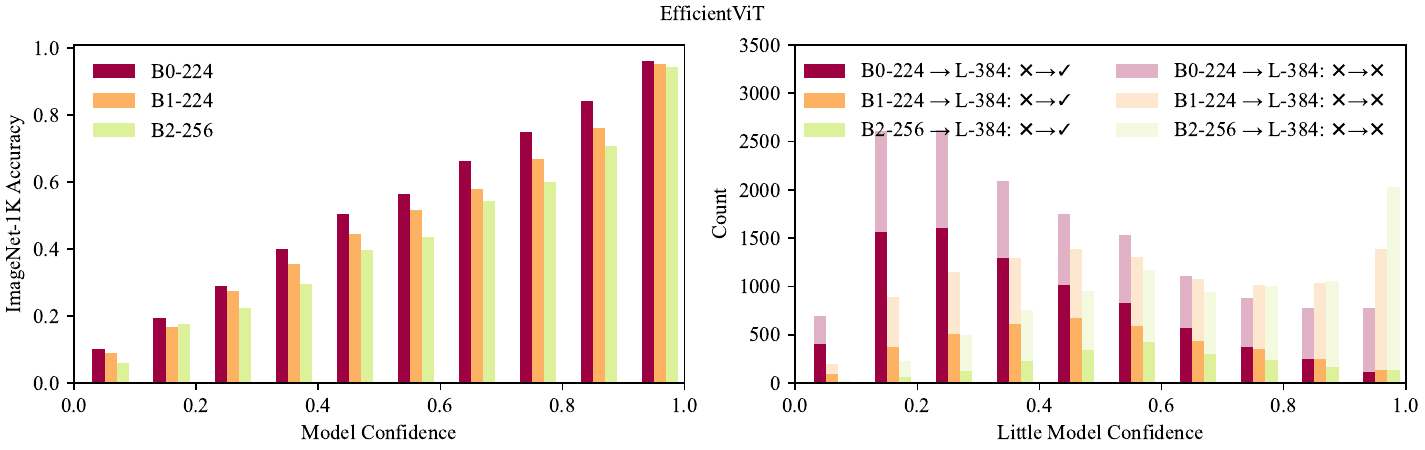}
          \\
          \includegraphics[width=0.9\linewidth]{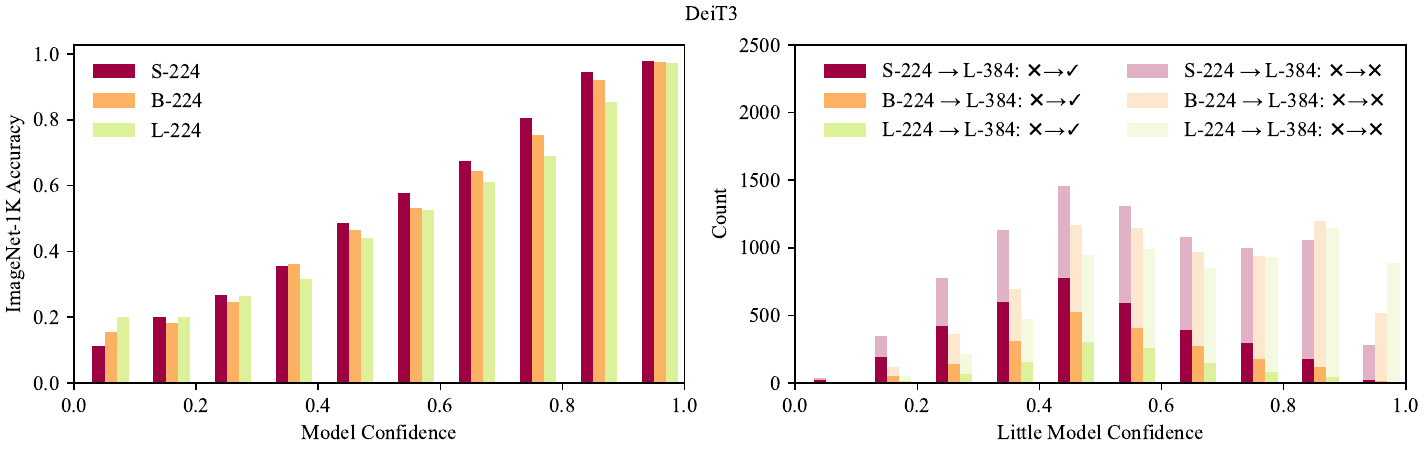}
          \\

    \end{tabular}
    \vspace{-1mm}
    \caption{More examples with ConvNext, EfficientViT, and DeiT3. Prediction confidence of individual models correspond well with prediction accuracy (left), which allows us to approximate a \textbf{``difficulty''} axis with prediction confidence. Breaking down the mistakes of little models by difficulty, we find that big models disproportionally ``correct'' mistakes that are difficult to the little models.
    }
    \label{fig:confidence_appendix}
    \vspace{-1mm}
\end{figure}

\begin{figure}[ht!]
    \centering

    \includegraphics[width=1.0\linewidth]{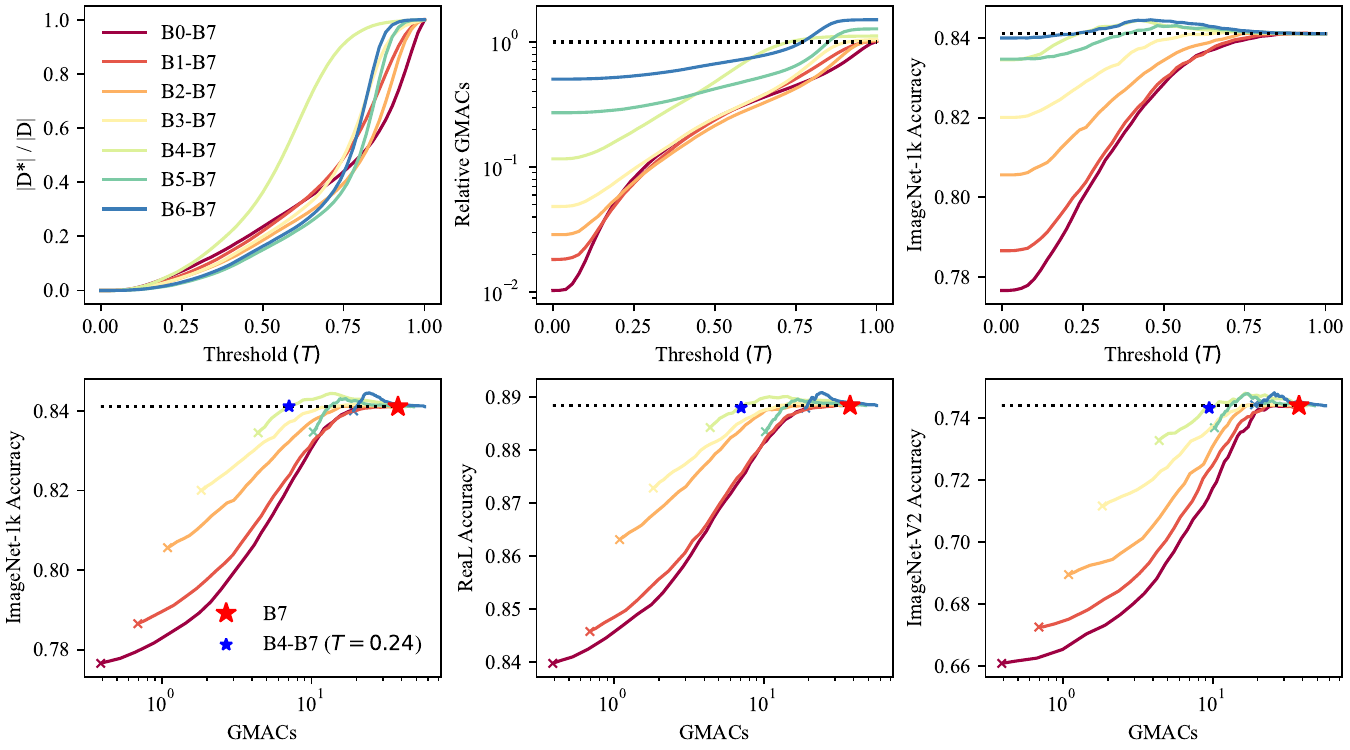}
    
    \caption{Speeding up EfficientNet-B7 with the EfficientNet family. Different from Figure \ref{fig:efficientnet_b7} where $T$ is chosen on ImageNet-1K, the optimal pair (blue star) is chosen on the smaller V2 set. This yields a similar optimal $T=0.28$ achieving 78\% of MACs reduction. 
    }
    \label{fig:efficientnet_b7_appendix}
    
\end{figure}

\end{document}